\let\footref\relax
\documentclass[11pt]{article}
\pdfoutput=1 

\usepackage{eamt23}
\usepackage{times}
\usepackage{url}
\usepackage{latexsym}
\usepackage[acronym]{glossaries-extra}
\usepackage[dvipsnames,table,xcdraw]{xcolor}
\usepackage[english=american]{csquotes}
\usepackage{enumitem}
\usepackage{subcaption}
\usepackage{booktabs}
\usepackage{makecell}
\usepackage{multicol}
\usepackage{multirow}
\usepackage{amsmath}
\usepackage{amssymb}
\usepackage{accents}

\makeatletter
\renewcommand\@biblabel[1]{}
\makeatother

\newcommand{\ccg}{\cellcolor{gray!25}{}}   

\newcommand{\ccwt}{\cellcolor{CadetBlue!15}{}} 
\newcommand{\ccgt}{\cellcolor{gray!35!CadetBlue!35}{}}

\newcolumntype{Y}{>{\centering\arraybackslash}X}
\newcolumntype{L}[1]{>{\raggedright\let\newline\\\arraybackslash\hspace{0pt}}m{#1}}
\newcolumntype{C}[1]{>{\centering\let\newline\\\arraybackslash\hspace{0pt}}m{#1}}
\newcolumntype{R}[1]{>{\raggedleft\let\newline\\\arraybackslash\hspace{0pt}}m{#1}}

\newcommand{\footlabel}[2]{%
    \addtocounter{footnote}{1}%
    \footnotetext[\thefootnote]{%
        \addtocounter{footnote}{-1}%
        \refstepcounter{footnote}\label{#1}%
        #2%
    }%
    $^{\ref{#1}}$%
}

\newcommand{\footref}[1]{%
    $^{\ref{#1}}$%
}

\title{Gender Lost In Translation: How Bridging The Gap Between Languages Affects Gender Bias in Zero-Shot Multilingual Translation}

\author{Lena Cabrera\textsuperscript{1}, Jan Niehues\textsuperscript{2}\\
  \textsuperscript{1}Department of Advanced Computing Sciences, Maastricht University, The Netherlands\\
  \textsuperscript{2}Institute for Anthropomatics and Robotics, Karlsruhe Institute of Technology, Germany\\
  {\tt l.cabreraperez@student.maastrichtuniversity.nl},\\ {\tt jan.niehues@kit.edu}}
  
\date{}

\newacronym{ai}{AI}{artificial intelligence}
\newacronym{ann}{ANN}{artificial neural network}
\newacronym{bpe}{BPE}{byte pair encoding}
\newacronym{cnn}{CNN}{convolutional neural network}
\newacronym{clm}{CLM}{conditional language model}
\newacronym{ce}{CE}{cross entropy}
\newacronym{grl}{GRL}{gradient reversal layer}
\newacronym{lm}{LM}{language model}
\newacronym{lstm}{LSTM}{long short-term memory}
\newacronym{ml}{ML}{machine learning}
\newacronym{mnmt}{MNMT}{multilingual NMT}
\newacronym{mt}{MT}{machine translation}
\newacronym{nlp}{NLP}{natural language processing}
\newacronym{nmt}{NMT}{neural machine translation}
\newacronym{rbmt}{RBMT}{rule-based machine translation}
\newacronym{rnn}{RNN}{recurrent neural network}
\newacronym{san}{SAN}{self-attention network}
\newacronym{sgd}{SGD}{stochastic gradient descent}
\newacronym{sota}{SOTA}{state of the art}
\newacronym{smt}{SMT}{statistical machine translation}
\newacronym{tm}{TM}{translation model}
\newacronym{oov}{OOV}{out of vocabulary}
\newacronym{zs}{ZS}{zero-shot}
\newacronym{pv}{PV}{pivoting}
\newacronym{wus}{WUS}{warmup steps}

\glsaddall

\begin{document}
\maketitle
\begin{abstract}
\Gls{nmt} models often suffer from gender biases that harm users and society at large.
In this work, we explore how bridging the gap between languages for which parallel data is not available affects gender bias in multilingual \gls{nmt}, specifically for zero-shot directions. 
We evaluate translation between grammatical gender languages which requires preserving the inherent gender information from the source in the target language.
We study the effect of encouraging language-agnostic hidden representations on models' ability to preserve gender and compare pivot-based and zero-shot translation regarding the influence of the bridge language (participating in all language pairs during training) on gender preservation.
We find that language-agnostic representations mitigate zero-shot models' masculine bias, and with increased levels of gender inflection in the bridge language, pivoting surpasses zero-shot translation regarding fairer gender preservation for speaker-related gender agreement.
\end{abstract}

\section{Introduction}
With the rapid proliferation of intelligent systems, machine learning models reflecting patterns of discriminatory behavior found in the training data is a growing concern of practitioners and academics.
Neural machine translation (NMT) models have proven notoriously gender-biased, often resulting in harmful gender stereotyping or an under-representation of the feminine gender in their outputs.
In recent years, several approaches to debias \gls{nmt} have been proposed, including debiasing the data before model training, the models during training, or post-processing their outputs.
However, to the best of the authors' knowledge, it has yet to be explored how the phenomenon of not observing enough data, if any, to model language accurately affects gender discrimination in \gls{mnmt}.

To support translation between language pairs never seen during training (i.e., zero-shot directions), two widely-used approaches leverage the language resources (i.e.,  parallel data) available during training: \textit{Pivot-based} translation uses an intermediate pivot/bridge language (as in source$\rightarrow$pivot$\rightarrow$target), whereas \textit{zero-shot} translation learns to bridge the gap between unseen language pairs using cross-lingual transfer learning.\footnote{We use \textquote{zero-shot \textit{directions}} to refer to language pairs unseen during training, whereas \textquote{zero-shot \textit{translation}} is \gls{nmt} capable of zero-shot inference, relying on a model’s generalizability to conditions unseen during training.}

In this work, we analyze gender bias in \gls{mnmt} in the context of \textit{gender preservation}, where gender information conveyed by the source language sentence needs to be preserved in the target language translation; in our experimental setting, source and target languages are grammatical gender languages that use a noun class system conforming with the \textit{gender binary}, i.e., the classification of gender into the opposite forms of feminine and masculine, considered indicative of a person's biological sex.\footnote{While gender, as opposed to biological sex, is viewed as a non-binary spectrum, many languages have not (yet) evolved beyond the male-female gender binary regarding linguistic gender when it ideally should correlate with biosocial gender.} 
We examine translations in terms of differences in gender preservation between both genders, which, if found, are evidence of gender-biased \gls{mt}.
More precisely, we focus on the impact that \textit{bridging the gap between unseen language pairs} has on the \gls{mt} models' ability to preserve the feminine and masculine gender, unambiguously indicated by the source sentence,
equally well in their outputs.
Our research questions are:
\begin{description}[labelwidth=22pt,leftmargin=!]
    \item[RQ1] {How do zero-shot and pivot-based translation compare regarding gender-biased outputs for zero-shot directions?}
    \item[RQ2] {Does the bridge language affect the gender biases perpetuated by zero-shot and pivot-based translations?}
    \item[RQ3] {Do translation quality improvements of zero-shot models reduce their gender biases?}
\end{description}

The remainder of this paper is structured as follows.
Section~\ref{sect:related_work} introduces the task of gender preservation in translation with relevant terminology and reviews related work on gender bias in \gls{nmt}.
Section~\ref{sect:methods} describes our experimental design, tailored toward investigating cause-and-effect relationships of gender bias in \gls{mnmt}.
Section~\ref{sect:data_evaluation} presents the data used and the evaluative procedure followed in our experiments.
Section~\ref{sect:results} presents the experimental setup and results, and Section~\ref{sect:conclusion} concludes with our summarized findings, limitations, and future research directions.

\section{Terminology \& Related Work}\label{sect:related_work}
In a large-scale analysis of the plethora of existing research addressing gender bias in \gls{nmt}, \newcite{savoldi2021gender} categorize them based on two conceptualizations of the problem: research works focusing on the weight of prejudice and stereotypes in \gls{nmt}, and studies assessing whether gender is preserved in translation.
In this paper, we analyze gender bias in \gls{mnmt} in the context of {gender preservation}, where for translation into a gender-sensitive target language, the gender information conveyed by the source language needs to be retained in the target language translation.

\paragraph{Gender in Lingustics:} In our gender bias evaluation we consider \textit{referential gender}, which, according to \newcite{cao2021toward}, only exists when an entity (i.e., a human) is mentioned and their gender (or sex) is realized linguistically.
Moreover, we focus on the translation between languages using \textit{grammatical gender}, a way of classifying nouns, assigning them gender categories (e.g., masculine, feminine, neuter, etc.) that may be independent of the real-world biosocial genders associated with referents; however, there is a tendency for languages to correlate grammatical gender with the gender of a referent, especially if human \cite{corbett1991gender,ackerman2019syntactic}.

For example, talking about a specific doctor (e.g., \textquote{the doctor loves $\text{\textit{her}}_F$ job}), the word choice of the female anaphoric pronoun is not determined by grammatical gender but only by referential gender.
The same sentence translated into German (\textquote{$\text{\textit{die}}_F$ $\text{\textit{Ärztin}}_F$ liebt $\text{\textit{ihren}}_F$ $\text{Job}_M$.}) requires the article (\textquote{die} = the) and pronoun (\textquote{ihren} = her) to agree with the feminine grammatical gender category the noun is assigned (\textquote{Ärztin} = female doctor).\footnote{Note, in German, the abstract noun \textquote{Job} is assigned the masculine grammatical gender category, while in English, \textquote{job} has no grammatical gender.}
On the other hand, the sentence \textquote{the doctor helps the nurse} without any further context information does not indicate the gender of either of the two mentioned entities; for the German translation, the gender of both the doctor (\textquote{$\text{{Arzt}}_M$}/\textquote{$\text{{Ärztin}}_F$}) and the nurse (\textquote{$\text{{Krankenpfleger}}_M$}/\textquote{$\text{{Krankenschwester}}_F$})~needs to be considered for the correct syntactic build-up of the sentence.
For details on the many differences in the manifestation of gender in languages, we refer the interested reader to related works such as that of \newcite{cao2021toward}.

\begin{figure*}[htb!]
    \begin{subfigure}{\textwidth}
        \centering
        \includegraphics[width=1\textwidth]{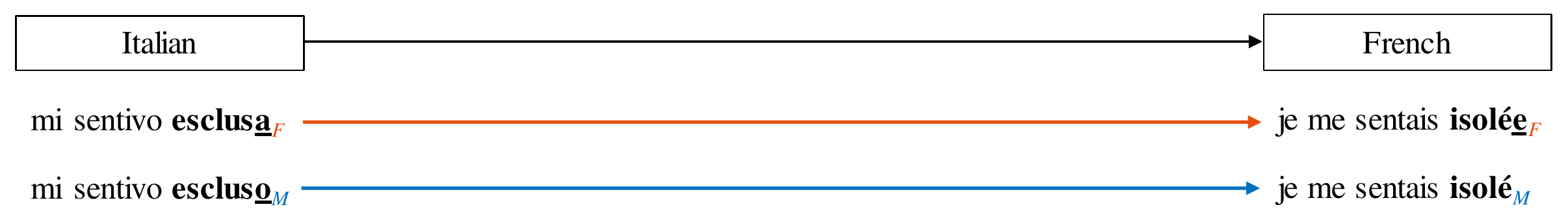}
        \caption{Illustration of the translation between grammatical gender languages (Italian$\leftrightarrow$French) examined in this work; here, for Italian$\rightarrow$French translation of the utterance \textquote{I felt alienated}. Information necessary to disambiguate gender (bold) was always conveyed by the source sentence (here, in Italian) and to be reflected in the translation (here, in French).}
        \label{fig:gender_preservation_explicit_cues}
    \end{subfigure}
    \hspace{0.5\textwidth}
    \begin{subfigure}{\textwidth}
        \centering
        \includegraphics[width=1\textwidth]{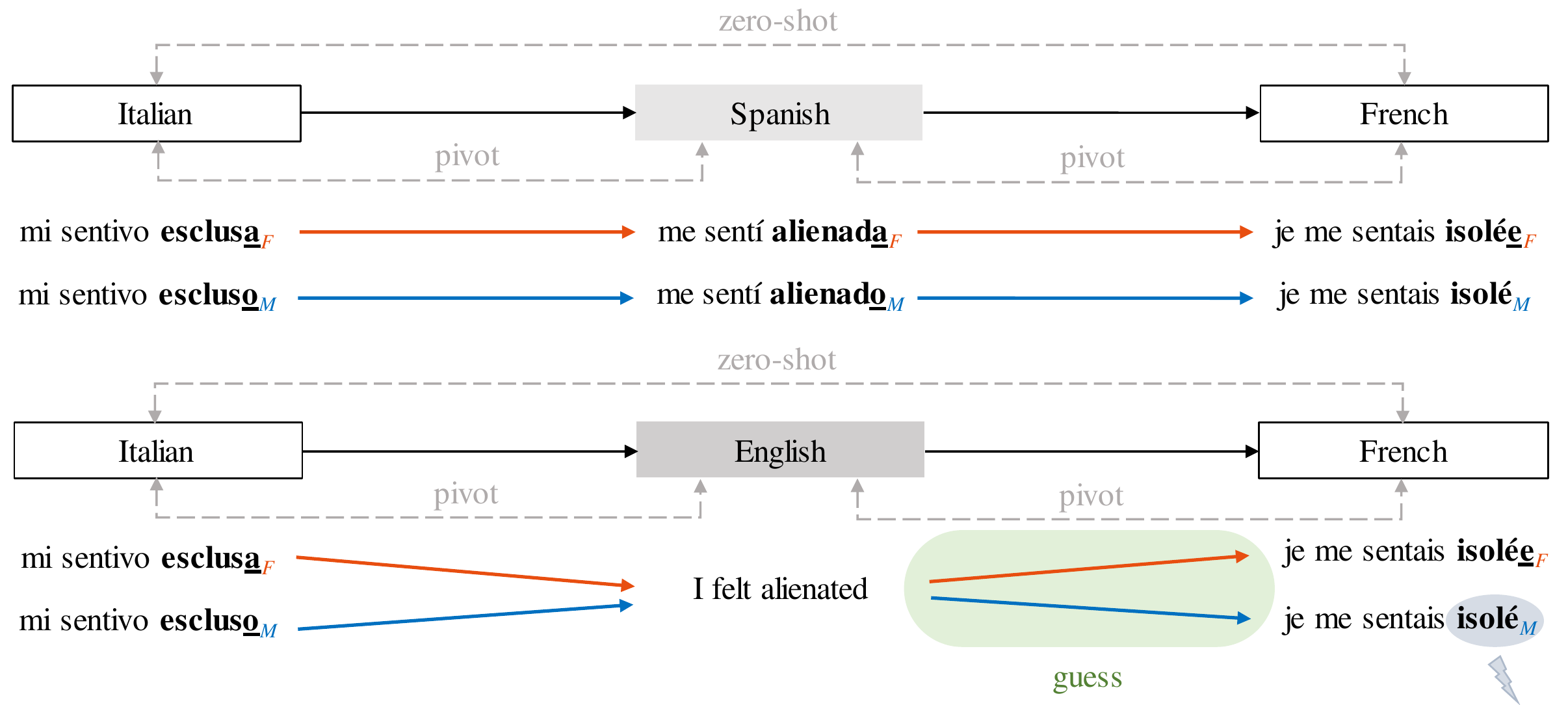}
        \caption{The richness of the gender-inflectional system of the bridge language, used to facilitate translation for unseen language pairs, affects models' ability to preserve the gender information from the source sentence.
        Scarcity of gender inflection in the bridge language (e.g., English) causes models to miss gender clues from the source and to resort to guessing the gender; when making the wrong guess, i.e., choosing the wrong gender as presented in the source, the model exhibits gender hallucination.}
        \label{fig:gender_preservation_bridging}
        \end{subfigure}
    \caption{Overview of our investigated translation scenario (here, for the utterance meaning \textquote{I felt alienated}): At inference, we translated between unseen gender-inflected source-target language pairs (i.e., Italian$\leftrightarrow$French) by bridging, implicitly (zero-shot) and explicitly (pivot-based), using bridge languages with different gender-inflectional systems (e.g., Spanish or English).}
    \label{fig:gender_preservation_}
\end{figure*}

\paragraph{Gender Preservation:} Translation into a gender-sensitive language, e.g., a grammatical gender language, involves gender agreement between nominal properties---e.g. grammatical and referential gender of a (pro)noun---and a determiner, adjective, verb, etc., depending on the target language agreement rules.
Whenever the source language is (largely) genderless, i.e., the gender of the noun is unspecified, and context information is unavailable, gender preservation is a non-trivial task for machines and humans alike.

In recent years, several approaches have been proposed to address the challenge of gender preservation. 
\newcite{vanmassenhove2018getting} leverage additional gender information by prepending a gender tag to each source sentence, both at training and inference time, to improve the generation of speakers' referential markings.
Avoiding the need for additional context information for training or inference, \newcite{basta2020towards} concatenate each sentence with its predecessor to achieve slight improvements in gender translation.
\newcite{moryossef2019filling} inject context information as they prepend a short phrase, e.g., \textquote{\textit{she} said to \textit{them}}, to the source sentence, translate the sentence with the prefix, and afterward remove the prefix translation from the model's output.
Specifying gender inflection in this way improves models' ability to generate feminine target forms, but it relies on (not always available) metadata about speakers and listeners.
Furthermore, different gender-specific translations in terms of word choices can be an arguably non-desirable side-effect.

A different approach is to post-process the output using counterfactual data augmentation.
\newcite{saunders2020reducing} use a lattice rescoring module that maps gender-marked words in the output to all possible inflectional variants and rescores all paths in the lattice corresponding to the different sentences with a model that has been gender-biased at the cost of lower translation quality.
Choosing the sentence with the highest score as the final translation results in increased accuracy of gender selection.
A downside is that data augmentation is very demanding for complex sentences with a variety of gender phenomena, such as those typically occurring in natural language scenarios.

\section{Analyzing Gender Bias in MNMT}\label{sect:methods}
In our experimental setting, information necessary to disambiguate gender was \textit{always} conveyed by the source sentence (cf. Figure~\ref{fig:gender_preservation_explicit_cues}) and, thus, available to the models.
Motivated by our research inquiry, we focused our investigation on the effect of bridging on gender preservation in \gls{mnmt} between unseen language pairs, as illustrated broadly in Figure~\ref{fig:gender_preservation_bridging}, exploring three influencing factors to learn about the cause-and-effect relationship of gender bias in \gls{mnmt}: \textit{i)}~the approach taken to bridge unseen language pairs (i.e., using {continuous} representations for zero-shot translation or {discrete} pivot language representations); \textit{ii)}~the choice of bridge language; and \textit{iii)}~language-agnostic model hidden representations.

\paragraph{Zero-Shot Translation Vs. Pivoting:}
To bridge the gap between an unknown source-target language pair at inference, we took two different approaches using the same trained translation model.
For \textit{pivot-based translation}, we cascaded a model to perform source$\rightarrow$pivot and pivot$\rightarrow$target translation. 
As such, pivoting used the pivot language as an explicit bridge between the unknown language pair.
For \textit{zero-shot translation}, we used the same model to translate directly between the unknown language pair, relying on the model's learned semantic space where sentences with the same meaning are mapped to similar regions regardless of the language.
Compared to pivoting, zero-shot translation circumvents error propagation and reduces computation time, but achieving high-quality zero-shot translations is challenging.
In light of our inquiry, we analyzed each approach's ability to preserve gender, comparing their performances for the feminine and the masculine gender.\footnote{In the presentation of our results, we use ZS and PV, short for zero-shot and pivot-based translation when space is limited.}

\paragraph{Bridge Language:}\label{methods:bridge_language}
English often participates in most, if not all, language pairs in a training corpus, making English, a language limited to pronominal gender (with a few exceptions), the most reasonable choice for a bridge language.
When translating into a genderless language (e.g., Hungarian), the potential loss of gender information conveyed by the source sentence is unproblematic as it is evidently without detrimental consequence.
However, when translating into a language with a \textit{higher} gender-inflected system than English (e.g., French or Italian), the loss of gender information poses a significant problem since the information necessary to disambiguate gender is virtually no longer existent (cf.~bottom in Figure~\ref{fig:gender_preservation_bridging}).

As preserving non-existent gender information is inherently impossible, also for humans, it is fair to assume that \gls{mt} models have difficulty when encountering this phenomenon of gender ambiguity; the simplest solution is to resort to \textit{random guessing}, with a 50\% chance of choosing one gender over the other.
Any other gender distribution ($\neq $~50:50\%) is not reflective of random guessing but instead indicative of \textit{educated guessing} based on knowledge or observations \textit{assumed} to be true that can, however, include biases.

Against this background, we studied the role of the bridge language in gender preservation, focusing on the gender bias differences between pivot-based and zero-shot translation, using bridge languages with different gender-inflectional systems, including English (low gender inflection), German and Spanish (high(er) gender inflection).
German and English are both Germanic languages.
Whereas in German, all noun classes require masculine, feminine, or neuter\footnote{In German, neuter gender inflection does not apply to nouns identifying people (cf. referential gender).} inflection, English lacks a similar grammatical gender system. 
In German, the gender of the noun is reflected in determiners like articles, possessives, and demonstratives. 
On the other hand, Spanish is a Romance language with a binary grammatical gender system, differentiating masculine and feminine nouns; from a grammatical point of view, there are no gender-neutral nouns.
The gender of nouns agrees with (some) determiners and, more often than in German, adjectives, making gender a pervasive feature in Spanish.

\paragraph{Language-Agnostic Hidden Representations:}\label{methods:model_modifications}
Since languages are characterized by different linguistic features, including those related to gender, it is reasonable to assume that language-\textit{specific} representations, tailored to the language pairs included during training, \textit{impair} gender preservation for unseen language pairs.
Because of this, we explored the effect of \underline{three modifications} to (the training of) a baseline Transformer \cite{vaswani2017attention} to encourage language-\textit{agnostic} hidden representations, which have proven to cause performance gains for zero-shot translation. 
We
\begin{itemize}
    \item \underline{removed a residual connection} in a middle Transformer encoder to \textit{lessen positional correspondences to the input tokens} and, thereby, reduced dependencies to language-specific word order ($R$) as proposed by \newcite{liu2021improving},
    \item encouraged \textit{similar (i.e., closer) source and target language representations} through an \underline{auxiliary loss} ($AUX$) similar to \newcite{pham2019improving} and \newcite{arivazhagan2019missing}, and
    \item performed \underline{joint adversarial training} \textit{penalizing recovery of source language signals} in the representations ($ADV$) as done by \newcite{arivazhagan2019missing}.
\end{itemize}
In our experiments, we examined the effect of these three modifications in isolation and tested some combinations; in total, we compared five different models to our baseline~($B$)---which we refer to as $B{+}AUX$, $B{+}ADV$, $R$, $R{+}AUX$, and $R{+}ADV$---to determine whether they mitigated models' gender biases.

\section{Evaluation Data \& Procedure}\label{sect:data_evaluation}
For our evaluation, we built on the work of \newcite{bentivogli2020gender} regarding the data and procedure used for our gender bias evaluation.

\subsection{Multilingual Gender Preservation Dataset}
In our experiments, we used the publicly available TED-based corpus MuST-C \cite{digangi2019mustc} for model training (cf.~Section~\ref{subsubsect:training_data} for details) and evaluated our models on a subset of MuST-SHE \cite{bentivogli2020gender}, a gender-annotated benchmark.
MuST-SHE is a subset of MuST-C and is available for English$\rightarrow$French, English$\rightarrow$Italian, and English$\rightarrow$Spanish translation, where at least one English gender-neutral word in a sentence needs to be translated into the corresponding masculine/feminine target word(s). 

The target languages included in MuST-SHE allowed us to investigate gender preservation for sentences where \textit{the source language always provides enough information to disambiguate gender}; with this research inquiry, two main criteria needed to be met by the evaluation data:
First, we wanted to evaluate gender translation \textit{between} grammatical gender languages.
Therefore, we formed a many-to-many subset from MuST-SHE, keeping only true-parallel data and realigning it to support evaluating translation between the three initial target languages.
Second, we wanted to investigate the gender biases in translation between language pairs unseen during training (i.e., zero-shot directions).
Using training corpora comprising different language pairs, we built models with different supervised translation directions. 
Accordingly, the models did not share the same zero-shot directions. 
For instance, a model trained on Spanish-X data had seen examples for language pairs that included Spanish.
Therefore, we discarded the Spanish examples and only used French-Italian examples in our evaluation to ensure equal zero-shot directions across all models considered in our experiments.

We obtained 278 sentences with detailed statistics presented in Table~\ref{tab:mustshe_statistics}.
The included French$\leftrightarrow$Italian directions left us with 556 translations for evaluation.
\begin{table}[htb!]
\centering
\resizebox{\columnwidth}{!}{
\begin{tabular}{lrrrrrr}
\toprule
& \multicolumn{2}{c}{\makecell{Feminine \\ (Female/Male)}} & \multicolumn{2}{c}{\makecell{Masculine \\ (Female/Male)}} & \multicolumn{2}{c}{\makecell{\textbf{Total} \\ (Female/Male)}} \\ \midrule
Cat. 1      & 64 & (64/0)   & 56 & (0/56)    & 120 & (64/56) \\
Cat. 2      & 72 & (58/14)  & 86 & (27/59)   & 158 & (85/73) \\
\cmidrule(lr){1-7}
\textbf{Total} & 136 & (122/14) & 142 & (27/115) & \textbf{278} & (149/129) \\ \bottomrule
\end{tabular}
}
\caption{Statistics of the \mbox{MuST-SHE} data used, broken down by referent gender (Feminine/Masculine), gender agreement (Cat.~1/2: speaker-related/speaker-independent), and speaker gender (Female/Male).}
\label{tab:mustshe_statistics}
\end{table}

The composition of this dataset, comprising French-Italian parallel data, provides different evaluative dimensions that can be considered for gender bias evaluation of \gls{mt} models. 

\paragraph{Referent Gender:}
Grammatical gender agreement determines the modification of certain words to express gender congruent with the other words they relate to, which, in our case, were the words designating a \textit{referent}---a person the speaker mentioned.
Consequently, the gender of a referent (cf. referential gender) determined the gender of gender-marked words relating to the referent (i.e., for a female referent, feminine inflected words, and for a male referent, masculine inflections).
All gender-marked words in a sentence did agree with the same (referent) gender.
As MuST-SHE is TED-based data, a referent was either the speaker, or a person not identified as the speaker (nor the addressee(s)/audience in our examples).

\paragraph{Speaker Gender:} 
Due to the evaluation data stemming from TED talks, examples are transcribed utterances spoken by different speakers of both feminine or masculine gender.
Depending on the type of gender agreement occurring in an utterance, the speaker's gender and referents' gender do or do not correlate.

\begin{figure*}[t!]
    \centering
    \includegraphics[width=0.9\textwidth]{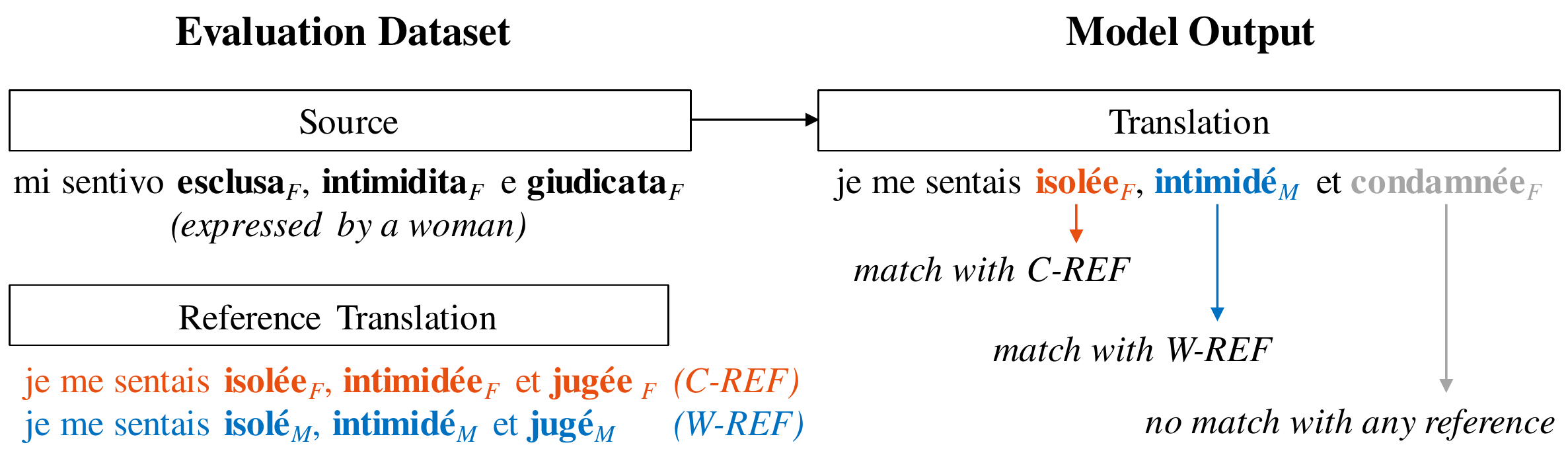}
    \caption{Illustration of the three possible translation outcomes of required gender preservation for Italian$\rightarrow$French translation of the utterance \textquote{I felt \textit{alienated}, \textit{intimidated}, and \textit{judged}}: The translation of a gender-inflected word either matched the correct reference translation \textit{C-REF} (here, \textquote{isolée} = alienated), the wrong reference translation \textit{W-REF} (here, \textquote{intimidé} = intimidated), or neither (here, \textquote{condamnée} = condemned).}
    \label{fig:correct_wrong}
\end{figure*}

\paragraph{Gender Agreement:}
Whenever the speaker was the referent, i.e., the speaker was referring to him- or herself, there is speaker-\textit{related} gender agreement among those gender-marked words referring to the speaker.
Languages with a less pronounced inflection of gender, such as English, can encounter syntactic structures that do not indicate a speaker's gender (cf.~bottom in Figure~\ref{fig:gender_preservation_bridging}).
In contrast, syntactic structures of languages with rich gender-inflected systems typically encode enough information to unambiguously classify a speaker's gender (cf.~top in Figure~\ref{fig:gender_preservation_bridging}).
Consequently, we hypothesized that using English as a bridge language results in the loss of gender information for sentences with speaker-related gender agreement; meanwhile, the higher gender-inflected grammatical gender languages, German and Spanish, were hypothesized to preserve the gender information when used as a bridge language. 

Whenever a person other than the speaker was the referent, i.e., the speaker was talking about someone else (e.g., \textquote{$\text{mi \textit{padre} se sentía alienado}_M$} = \textquote{my dad felt alienated} uttered by a \textit{female} speaker), there is speaker-\textit{independent} gender agreement among those gender-marked words referring to the referent. 
For these examples in our data, meaning construction typically does not require the integration of semantic information about the speaker for correct syntactic processing and translation.
The gender inflection of words is therefore often purely based on syntactic agreement with a formally marked subject (here, the referent), making the referent's gender identity explicit in those utterances for all three considered bridge languages, English, German, and Spanish.

\subsection{Method of Measurement}
Similar to \newcite{bentivogli2020gender}, we used the concept of gender-swapping to measure how often a model preserved the gender compared to how often it produced the opposite gender form, thus opting for the wrong instead of the correct gender, which, if frequently done, signaled models' acting on gender biases.

Following this idea, models' generated translations of gender-marked words belonged to one of three categories, which we exemplify using Figure~\ref{fig:correct_wrong}.
First, the \textit{expected translation}, for which we measured how often the \textit{correct} translation (ground truth)---specified by a reference translation C-REF---was produced (e.g., \textquote{isolée} in the exemplary model output in Figure~\ref{fig:correct_wrong}).
Second, the \textit{gender-reversed translation}, for which we measured how often the translation was \textit{wrong}, but only regarding the gender inflection of gender-marked words---specified by a reference W-REF---i.e., instead of the required correct gender realization as per ground truth (e.g., the feminine adjective \textquote{intimidé\underline{e}}), the model produced the opposite gender form (e.g., the masculine adjective \textquote{intimidé}). 
Third, a \textit{translation different from both reference translations}, e.g., instead of \textquote{jugée} (C-REF) or \textquote{jugé} (W-REF), the model produced the adjective \textquote{condamnée}, or any other word not matching C-REF or W-REF; in this case, we had no reference as to whether the gender inflection, regardless of the predicted word base, was correct or wrong, forcing us to exclude these translations from our gender bias evaluation.

We used two metrics to evaluate our models: BLEU (similar to \newcite{bentivogli2020gender}) and accuracy.
For the accuracy on feminine and masculine word forms, we measured how often a model was able to produce the correct gender ($C$) for those words that matched either the correct or the wrong reference set ($C{+}W$); we refer to this as \textit{gender preservation} ($\alpha_{\mathrm{correct}}$).
As we only relied on correct and wrong \textquote{matches} ($C{+}W$)---excluding words that did not match any reference set ($N$)---the larger in size this set was, i.e., the larger the sample size, the more significant our findings; therefore, we weighted $\alpha_{\mathrm{correct}}$ by the size of $C{+}W$ in relation to the number of all translations ($C{+}W{+}N$), matching a reference ($C{+}W$) or not matching any reference ($N$); we refer to this weighting factor as \textit{sample size} ($\rho$). 
Formally, we defined the accuracy~$\gamma$ to measure the \textit{gender preservation performance weighted by the sample size} as follows:
\begin{align*}
  \gamma = 
    \underbrace{\frac{C}{C+W}}_{\alpha_{\mathrm{correct}}} \cdot \underbrace{\frac{C+W}{C+W+N}}_{\rho} = \frac{C}{C+W+N}
\end{align*}

To compare the performances for the two genders, we computed the \textit{gender gap} $\delta$ between results for feminine and the masculine word forms:
\begin{align*}
    \delta = 1 - \frac{\mathrm{min}(\gamma^{\mathrm{F}}, \gamma^{\mathrm{M}})}{\mathrm{max}(\gamma^{\mathrm{F}}, \gamma^{\mathrm{M}})}
    \label{eq:gender_gap}
\end{align*}
As a reflection of gender biases, gender gaps should be as small as possible and ideally zero due to minimal differences between the results for the feminine and the masculine gender.
Furthermore, we analyzed the difference between scores for the correct and the wrong references to determine whether translations were gender-biased.

\section{Experiments \& Results}\label{sect:results}
The code and scripts used for our experimental evaluation are available on GitHub.\footnote{\url{https://github.com/lenacabrera/gb_mnmt}}

\subsection{Experimental Setup}

\paragraph{Training Data:}\label{subsubsect:training_data}
In our experiments, we used the publicly available corpora MuST-C \cite{digangi2019mustc} for model training.
To investigate the impact of the bridge language, determined by the language pairs included during training, we formed three training corpora that are subsets of MuST-C (X),\footnote{From release version 1.2, we included 10 of the 15 available languages: Czech, Dutch, English, French, German, Italian, Portuguese, Romanian, Russian, and Spanish.} with language pairs en$\leftrightarrow$X$\backslash$en, de$\leftrightarrow$X$\backslash$de, and es$\leftrightarrow$X$\backslash$es, where X$\backslash$en is the language set X excluding English~(en), German~(de), or Spanish~(es). 
On each of the three corpora, we trained a model and afterward evaluated the three trained models on our evaluation data.
Since only a portion (\raisebox{-0.9ex}{\~{}}10\%) of MuST-C is true-parallel data, the training corpora differed in size, as specified in Table \ref{tab:mustc_statistics}.
\begin{table}[hbt!]
\centering
\begin{tabular}{lr}
\toprule
Language Pairs & \# Sentences per Direction \\
\midrule
en~$\leftrightarrow$~X$\backslash$en & 125,000--267,000 \\
de~$\leftrightarrow$~X$\backslash$de & 103,000--223,000 \\
es~$\leftrightarrow$~X$\backslash$es & 102,000--258,000 \\
\bottomrule
\end{tabular}
\caption{Overview of the three MuST-C subsets used.}
\label{tab:mustc_statistics}
\end{table}

\paragraph{Preprocessing:}
MuST-C comes with partitioned training and validation sets which we kept unchanged in our experiments, except for the modifications described above.
For the training and validation data, we first performed tokenization and truecasing using the Moses\footnote{\url{https://github.com/moses-smt/mosesdecoder}} tokenizer and truecaser.
Afterward, we learned \gls{bpe} using subword-nmt\footnote{\url{https://github.com/rsennrich/subword-nmt}} \cite{sennrich2015neural}.
We performed 20 thousand merge operations and only used tokens occurring in the training set with a minimum frequency of 50 times.
Our evaluation data was preprocessed in a similar way using the \gls{bpe}-learned vocabulary.

\paragraph{Training \& Inference Details:}\label{sect:experimental_setup__training}
Our baseline~($B$) was a Transformer with 5 encoder and 5 decoder layers with 8 attention heads, an embedding size of 512, and an inner size of 2048.
For regularization, we used dropout with a rate of 0.2 and performed label smoothing with a rate of 0.1.
Moreover, we used the learning rate schedule from \newcite{vaswani2017attention} with 8,000 \gls{wus}.
The source and target word embeddings were shared.
To specify the output language, we used a target-language-specific beginning-of-sentence token.
As part of our model modifications, we removed a residual connection~($R$) in the third encoder layer \cite{liu2021improving}.
We trained each model for 64 epochs and averaged the weights of the five best checkpoints ordered by the validation loss.
For the auxiliary similarity loss~($AUX$) and the adversarial language classifier~($ADV$), we resumed training of the baseline and the model with removed residual connections for 10 additional epochs (400 \gls{wus}).
By default, we only included supervised directions in the validation set.
To compute BLEU scores, we used sacreBLEU \cite{post2018call}, which provides a fair and reproducible evaluation, as it operates on detokenized text.

\subsection{Results}
\begin{figure}[ht!]
    \centering
    \includegraphics[width=1\columnwidth]{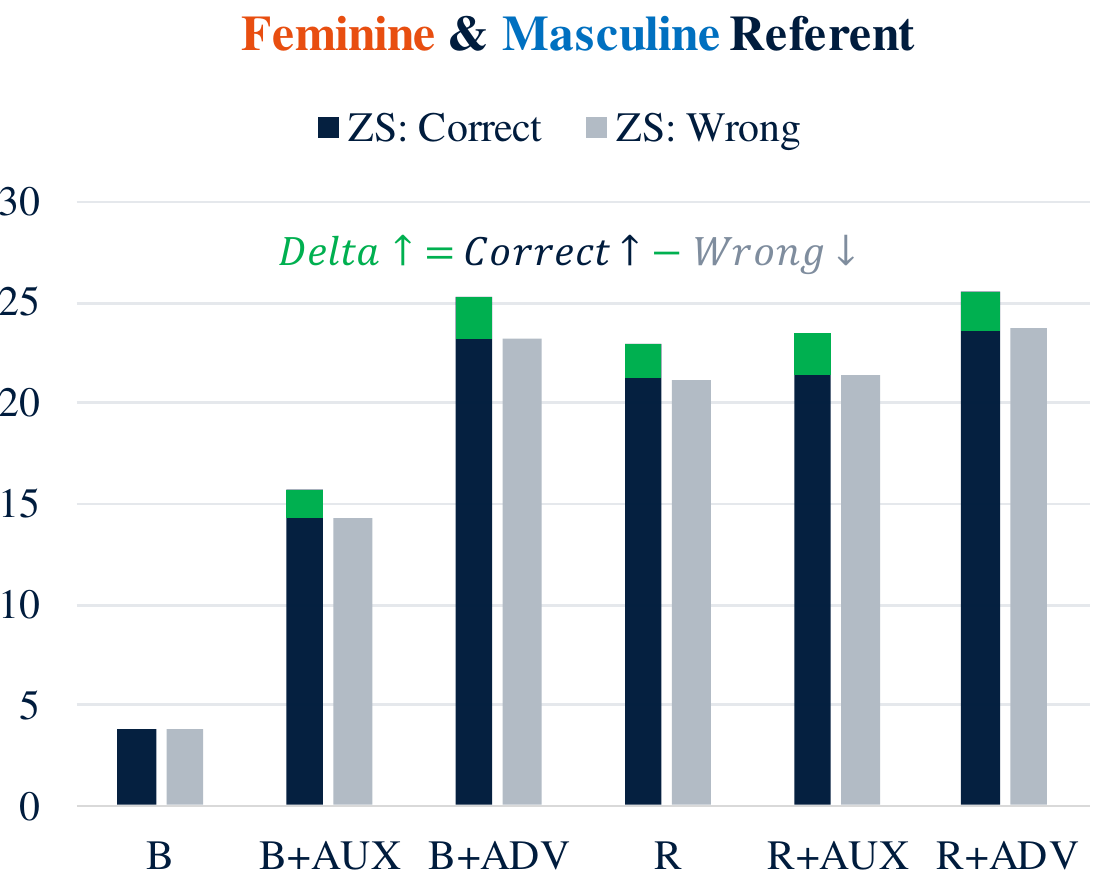}
    \caption{Average BLEU scores for  \textit{Correct} (left bar, higher $\uparrow$ is better) and \textit{Wrong} (right bar, lower $\downarrow$ is better) MuST-SHE references of our six evaluated zero-shot models, complemented with the delta (green bar, higher $\uparrow$ is better) between both. Results are for the feminine and masculine referent gender.\footref{fn:en_zs_bridge_models}}
    \label{fig:mod_bleu_zs_6_models}
\end{figure}

In Figure~\ref{fig:mod_bleu_zs_6_models}, we present the BLEU scores indicative of the similarity of the generated translations of \mbox{MuST-SHE} utterances to the \textit{Correct} references and their gender-reversed counterparts (\textit{Wrong} references) regardless of the referent gender, as well as the difference (delta) between \textit{Correct} and \textit{Wrong} scores for zero-shot models only.\footlabel{fn:en_zs_bridge_models}{Results are for models trained on en$\leftrightarrow$X$\backslash$en data.}

The bar graph illustrates that modifying our baseline $B$ to encourage language-agnostic representations improves the poor gender preservation performance of $B$ noticeably when performing zero-shot translation.
While the delta between \textit{Correct} and \textit{Wrong} scores for $B$ is zero, we consistently observe positive deltas (cf. green bars) that signal more correct than wrong gender translations; hence, through more language-agnostic hidden representations the modified zero-shot models more often can recover information (conveyed by the source language sentence) necessary to preserve the gender in the target language translation which, in turn, reduces the number of translations produced based on reflecting learned gender biases (in response to RQ3).
It shows that $R{+}ADV$, closely followed by $B{+}ADV$, yields the highest \textit{Correct} BLEU scores (higher is better) and one of the largest deltas between \textit{Correct} and \textit{Wrong} scores (higher is better); therefore, we take a closer look at the performance of $R{+}ADV$. 

\begin{figure}[ht!]
    \centering
    \includegraphics[width=1\columnwidth]{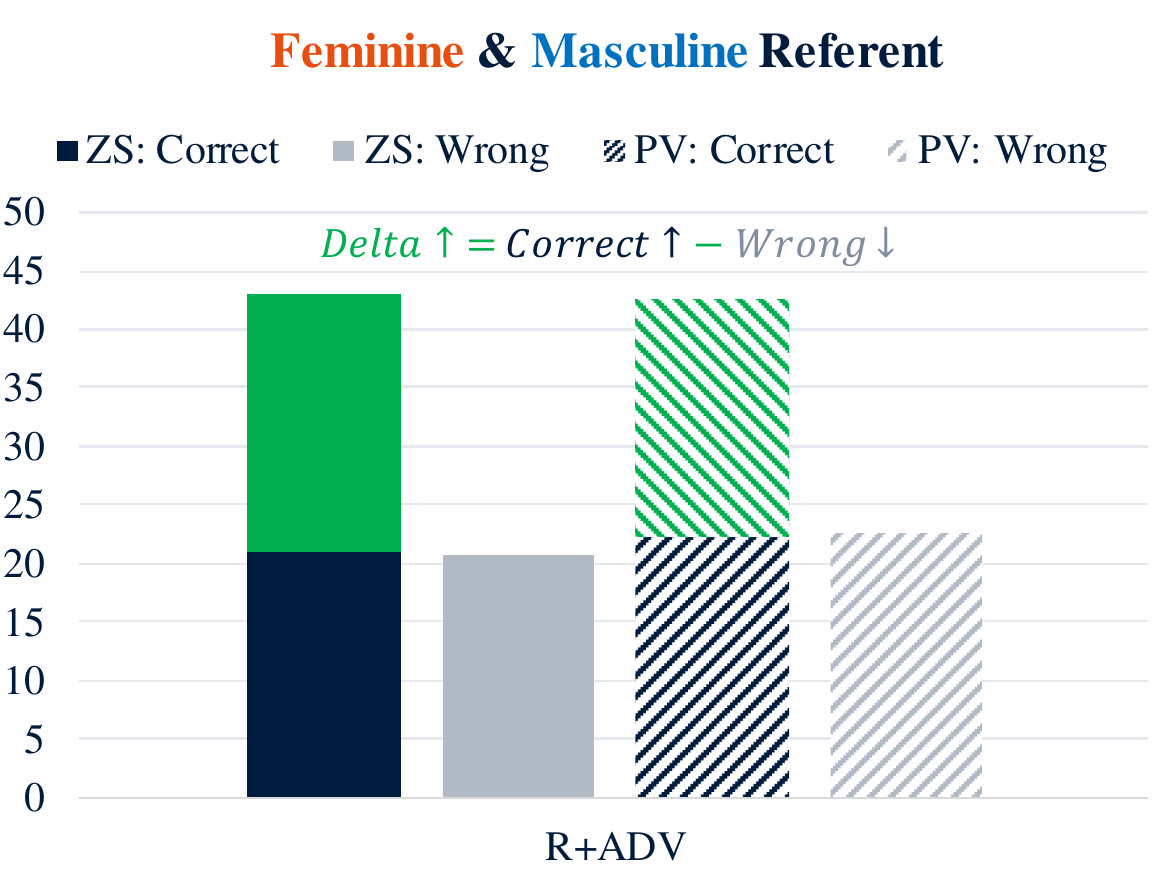}
    \caption{Average accuracy scores of zero-shot translation (full bars) and pivoting (hatched) for \textit{Correct} (left bar, higher $\uparrow$ is better) and \textit{Wrong} (right bar, lower $\downarrow$ is better) MuST-SHE references complemented with the delta (green bars, higher $\uparrow$ is better) between both for the model $R{+}ADV$. Results are for the feminine and masculine referent gender.\footref{fn:en_zs_bridge_models}}
    \label{fig:rq2_zs_pv_radv}
\end{figure}

Complementary to the BLEU-based evaluation, we examine $R{+}ADV$ accuracies ($\gamma$), where better or worse performance measured is reliably attributed to better or worse translation of \textit{gender-inflected words only}.
From Figure~\ref{fig:rq2_zs_pv_radv}, we can observe very similar performances for zero-shot and pivot-based translation using $R{+}ADV$ (RQ1).

While both approaches achieve similar \textit{Correct} accuracy scores (43.0 for ZS and 42.5 for PV), we observe slightly lower \textit{Wrong} scores for zero-shot translation (20.8) than for pivoting (22.5).
As a result, the delta for zero-shot is higher (better) than for pivot-based translation (22.2 vs. 20.2).

\begin{figure*}[t!]
    \centering
    \includegraphics[width=1\textwidth]{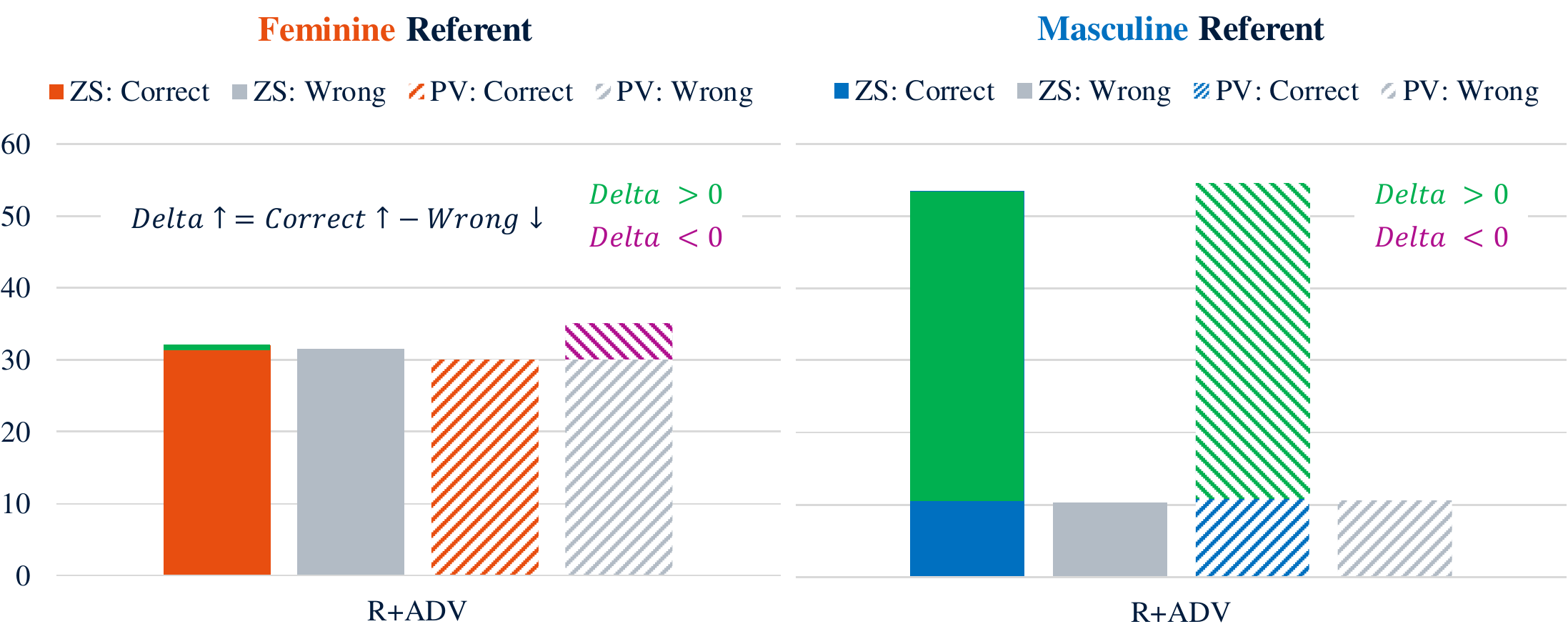}
    \caption{Average accuracy scores of zero-shot translation (full bars) and pivoting (hatched) for \textit{Correct} (left bar, higher $\uparrow$ is better) and \textit{Wrong} (right bar, lower $\downarrow$ is better) MuST-SHE references, complemented with the delta (green [\textit{Delta} $>0$] and magenta [\textit{Delta} $<0$] bars, higher $\uparrow$ is better) between both for the model $R{+}ADV$. Results are broken down by referent gender (feminine [left] vs. masculine [right]).\footref{fn:en_zs_bridge_models}}
    \label{fig:rq2_zs_pv_radv_f_m}
\end{figure*}

To gain better insight into the difference in gender preservation between both approaches, we break down the accuracies and compare them for the feminine and masculine gender; the corresponding results are depicted in Figure~\ref{fig:rq2_zs_pv_radv_f_m}.
The large differences between the accuracies for feminine and masculine referents clearly show that the model is acting according to a \textit{masculine bias} that detriments feminine and benefits masculine preservation of gender signals conveyed by the source sentence.
The \textit{Correct} accuracies in the masculine case are almost twice as high as their feminine counterparts.
Furthermore, comparing the \textit{Wrong} accuracies, we see an even bigger difference, as masculine \textit{Wrong} scores are much smaller (by a factor of 5), whereas feminine \textit{Wrong} scores are almost identical to their \textit{Correct} counterparts.

In the masculine case, performances by both approaches are very similar, with pivoting achieving slightly higher \textit{Correct} and \textit{Wrong} scores (54.5 vs. 53.4 and 10.6 vs. 10.4).
In the feminine case, we see that zero-shot translation is more accurate regarding feminine gender preservation: The delta between \textit{Correct} and \textit{Wrong} accuracies is small but positive (0.5), whereas for pivoting, we observe a negative delta (-4.9) that signals more wrong (masculine) than correct (feminine) translations for words where the required gender realization is feminine.
Accordingly, it turns out that zero-shot translation performs noticeably better for feminine gender preservation---which is generally poorer than masculine gender preservation---compared to pivoting and, as a consequence, mitigates the masculine bias to a larger extent, producing more balanced gender outputs (RQ1).

As we assumed the bridge language to play an important role in gender preservation, we compare the model's performance for zero-shot and pivot-based translation when trained using different training corpora that enabled the use of different bridge languages, namely English (for the results presented so far) and the grammatical gender languages German and Spanish (in response to RQ2).
As we expected to see differences between the three languages regarding sentences with and without speaker-related gender agreement, we present the \textit{Correct} accuracies broken down by referent gender and complemented with the gender gap ($\delta$) between feminine and masculine accuracies for either utterance category in Table~\ref{tab:bridging_results_gender_gap}.
\begin{table}[htb!]
\centering
\resizebox{\columnwidth}{!}{
\begin{tabular}{lccccC{10mm}C{10mm}}
\toprule
  \multicolumn{1}{l}{\multirow{2}[2]{*}{\makecell[l]{Bridge\\Language}}} &
  \multicolumn{2}{c}{\makecell{{Feminine} $\uparrow$}} &
  \multicolumn{2}{c}{\makecell{{Masculine} $\uparrow$}} & 
  \multicolumn{2}{c}{\ccwt{\textbf{Gender Gap} $\downarrow$}}
  \\ 
  
\cmidrule(r){2-3} \cmidrule{4-5} \cmidrule{6-7} 
    \multicolumn{1}{l}{} &
    \multicolumn{1}{c}{ZS} & \multicolumn{1}{c}{PV} &
    \multicolumn{1}{c}{ZS} & \multicolumn{1}{c}{PV} &
    \multicolumn{1}{c}{\ccwt{ZS}} & \multicolumn{1}{c}{\ccwt{PV}} \\ 
\cmidrule{1-7}\morecmidrules\cmidrule{1-7}
& \multicolumn{6}{c}{Speaker-\textit{Independent} Gender Agreement} \\ 
\cmidrule{1-7}
\ccg{English} & \ccg{\underline{42.8}} & \ccg{39.8} & \ccg{56.7} & \ccg{\underline{\textbf{58.3}}} & \ccgt{\underline{{0.25}}}  & \ccgt{0.32} \\
\multicolumn{1}{l}{German} & 40.4 & \underline{43.6} & 50.1 & \underline{55.6} & \ccwt{\underline{{0.19}}} & \ccwt{0.22} \\
\multicolumn{1}{l}{Spanish} & \underline{\textbf{49.6}} & 45.3 & \underline{57.7} & 55.0 & \ccwt{\underline{\textbf{0.14}}} & \ccwt{0.18} \\
\cmidrule{1-7}\morecmidrules\cmidrule{1-7}
& \multicolumn{6}{c}{Speaker-\textit{Related} Gender Agreement} \\ 
\cmidrule{1-7}
\ccg{English} & \ccg{\underline{20.2}} & \ccg{19.2} & \ccg{48.2} & \ccg{\underline{48.7}} & \ccgt{\underline{{0.58}}} & \ccgt{0.61} \\
\multicolumn{1}{l}{German} & 15.1 & \underline{18.4} & \underline{\textbf{51.1}} & 49.8 & \ccwt{0.70}  & \ccwt{\underline{{0.63}}} \\
\multicolumn{1}{l}{Spanish} & 23.8 & \underline{\textbf{29.4}} & \underline{50.6} & 45.7 & \ccwt{0.53}  & \ccwt{\underline{\textbf{0.36}}} \\

\bottomrule
\end{tabular}
}
\caption{Average accuracy scores for \textit{Correct} (higher $\uparrow$ is better) references with speaker-related and speaker-independent gender agreement when bridging via English, German or Spanish using the model $R{+}ADV$. Results are broken down by referent gender and complemented with the gender gap (lower $\downarrow$ is better) between feminine and masculine \mbox{accuracies}. Underlined scores are the best of both approaches, and bold scores are the best across languages.}
\label{tab:bridging_results_gender_gap}
\end{table}

It shows that the performances for speaker-independent gender agreement are noticeably better (i.e., higher accuracies and smaller gender gaps) than for speaker-related gender agreement, which can be attributed to reduced gender ambiguity due to more explicit gender clues provided by source sentences in the former case.
It shows that the poorer performance for speaker-related gender agreement affects the feminine gender more than the masculine gender when considering the much smaller difference in results for masculine word forms compared to a significant drop in scores for feminine word forms for speaker-related gender agreement (again, this very prominently highlights the model's masculine bias).
Consequently, it shows that the feminine discrimination found throughout all models' performances is more prominent in cases of high gender ambiguity, confirming the notion of models making \textquote{educated} gender guesses that are tainted by gender biases. 

Moreover, our results reveal clear differences in gender preservation between languages for both types of gender agreement: 
For speaker-independent gender agreement (e.g., \textquote{$\text{mi \textit{padre} se sentía alienado}_M$} = \textquote{my dad felt alienated}), we find that zero-shot translation produces smaller gender gaps compared to pivoting for all three bridge languages.
For the English bridge, the difference between zero-shot translation and pivoting is most pronounced, albeit small.
For speaker-related gender agreement (e.g., \textquote{me sentí $\text{alienada}_F$} = \textquote{I felt alienated}), it turns out that zero-shot translation achieves a slightly smaller gender gap compared to pivoting using the English bridge language (where gender information is likely lost); for the German and the Spanish bridge languages, we observe better pivoting results regarding smaller gender gaps and, thus, more balanced correct gender outputs.
This outcome confirms our hypothesis that for languages where gender inflection is relatively low, zero-shot translation is not as much affected by a loss of gender information (which impairs gender preservation for pivoting using discrete language representations), as it relies on more language-agnostic gender clues likely found in the continuous representations.
Moreover, the outcomes suggest that with an increased level of gender inflection in the bridge language, pivoting surpasses zero-shot translation regarding fairly balanced gender preservation for speaker-related gender agreement.

\section{Conclusion}\label{sect:conclusion}
In this paper, we explored gender bias in \gls{mnmt} in the context of gender preservation for zero-shot translation directions, i.e., unseen language pairs (French$\leftrightarrow$Italian), compared the performances of pivoting and zero-shot translation using discrete and continuous representations respectively, studied the influence the bridge language has on both approaches, and examined the effect language-agnostic representations have on zero-shot models' gender biases.
Based on our experimental results, we addressed three research questions.

\begin{description}[labelwidth=22pt,leftmargin=!]
    \item[RQ1] {How do zero-shot and pivot-based translation compare regarding gender-biased outputs for zero-shot directions?}
\end{description}
We find that zero-shot translation and pivoting achieve similar gender preservation performances, but zero-shot translation better preserves the feminine gender, which mitigates the masculine bias---the consistently worse feminine than masculine results across all evaluated models and both approaches---more than pivoting when bridging via English.

\begin{description}[labelwidth=22pt,leftmargin=!]  
    \item[RQ2] {Does the bridge language affect the gender biases perpetuated by zero-shot and pivot-based translations?}
\end{description}
Our experiments revealed that the bridge language affects gender biases in \gls{mnmt}.
For English, a language limited to pronominal gender (with a few exceptions), we find that zero-shot translation performs better than pivoting regarding a more fairly balanced preservation of feminine and masculine gender.
Using two richer gender-inflected bridge languages, Spanish and German, revealed that with an increased level of gender inflection in the bridge language, pivoting surpasses zero-shot translation regarding fewer gender-biased outputs for utterances with speaker-related gender agreement.

\begin{description}[labelwidth=22pt,leftmargin=!]  
    \item[RQ3] {Do translation quality improvements of zero-shot models reduce their gender biases?}
\end{description}

All three evaluated modifications encouraging language-agnostic hidden representations (cf.~Section~\ref{methods:model_modifications}) improved zero-shot models' ability to preserve the feminine and masculine gender and reduced the gap between better masculine and worse feminine results; they improved zero-shot models' performances to the point where they outperformed pivoting regarding more fairly balanced preservation of both genders when bridging via English.

Besides our findings, this work also features some limitations that can be addressed in future work.
First, the data used in our experimental evaluation limited the scenarios to those examined.
Future work can examine the translation of sentences with mixed gender (i.e., sentences including feminine \textit{and} masculine word forms) and directions, including languages from different language families and with different gender systems, to further study language differences.
Second, developing a large-scale gender-annotated corpus suitable for \gls{mnmt} training could most likely be used to improve models' gender preservation performance.
A well-performing gender classifier could be used to annotate the \mbox{MuST-C} dataset with token- or word-level gender labels.
Third, we believe that the metrics currently used to evaluate models' gender biases are not ideal.
For instance, model outputs mismatching the reference translations used for evaluation are discarded, despite potentially being appropriate translations (e.g., synonyms); future work could explore using additional morphological analysis tools to include those translations in the gender bias evaluation.
Generally, inquiring about the phenomenon of gender bias in translation requires appropriate and established metrics; the lack thereof currently leaves room for improvement in evaluative procedures.

While there is a lot of potential for further research on this topic, it is crucial to acknowledge that, ultimately, translation technology is bound by the principles of language, which subtly reproduces societal asymmetries and embeds signs of sexism, including masculine defaults and more subtle conventions by which expressions referring to females are grammatically more complex in many languages.
Consequently, combating gender biases in translation technology requires awareness of language use, as it is one of the most powerful means through which sexism and gender discrimination are perpetrated and reproduced.

\bibliography{eamt23}
\bibliographystyle{eamt23}
\end{document}